\documentclass[sn-mathphys]{sn-jnl}

\jyear{2021}%

\theoremstyle{thmstyleone}%
\usepackage{xcolor}
\newcommand{\ours}{FIVRS}

\theoremstyle{thmstyletwo}%

\usepackage{subfigure}

\theoremstyle{thmstylethree}%

\begin{document}

\title[Fully Immersive Virtual Reality for Skull-base Surgery]{Fully Immersive Virtual Reality for Skull-base Surgery: Surgical Training and Beyond}


\author*[1]{\fnm{Adnan} \sur{Munawar}}\email{amunawa2@jh.edu}

\author[1]{\fnm{Zhaoshuo} \sur{Li}}

\author[2]{\fnm{Nimesh} \sur{Nagururu}}

\author[2]{\fnm{Danielle} \sur{Trakimas}}

\author[1]{\fnm{Peter} \sur{Kazanzides}}

\author[1]{\fnm{Russell H.} \sur{Taylor}}

\author[2]{\fnm{Francis X.} \sur{Creighton}}

\affil[1]{\orgdiv{Johns Hopkins University}, \orgaddress{\city{Baltimore}, \postcode{21218}, \state{MD}, \country{USA}}}

\affil[2]{\orgdiv{Johns Hopkins Medicine}, \orgaddress{\city{Baltimore}, \postcode{21209}, \state{MD}, \country{USA}}}

\abstract{
\textbf{Purpose:}
A virtual reality (VR) system, where surgeons can practice procedures on virtual anatomies, is a scalable and cost-effective alternative to cadaveric training. 
The fully digitized virtual surgeries can also be used to assess the surgeon's skills using measurements that are otherwise hard to collect in reality.
Thus, we present the Fully Immersive Virtual Reality System (FIVRS) for skull-base surgery, which combines surgical simulation software with a high-fidelity hardware setup.

\textbf{Methods:} \ours{} allows surgeons to follow normal clinical workflows inside the VR environment. 
\ours{} uses advanced rendering designs and drilling algorithms for realistic bone ablation. 
A head-mounted display with ergonomics similar to that of surgical microscopes is used to improve immersiveness.
Extensive multi-modal data is recorded for post-analysis, including eye gaze, motion, force, and video of the surgery. 
A user-friendly interface is also designed to ease the learning curve of using \ours.

\textbf{Results: }
We present results from a user study involving surgeons with various levels of expertise. The preliminary data recorded by \ours{} differentiates between participants with different levels of expertise, promising future research on automatic skill assessment. Furthermore, informal feedback from the study participants about the system's intuitiveness and immersiveness was positive.

\textbf{Conclusion:}
We present \ours, a fully immersive VR system for skull-base surgery. 
\ours{} features a realistic software simulation coupled with modern hardware for improved realism. The system is completely open-source and provides feature-rich data in an industry-standard format.
}




\maketitle

\section{Introduction}
\label{sec: Introduction}
Skull-base surgery requires surgeons to operate within the skull to remove bone and soft tissue for various procedures. These procedures are among the most complicated surgical interventions~\cite{scholz2010skull} as operating within the skull requires safely navigating around sensitive structures including nerves and vessels that are often concealed by operable tissue sub-millimeter distances away. Furthermore, there can be significant anatomic variation between patients~\cite{cousins2008lateral}. The above challenges require high expertise by the surgeon, gained through extensive training and practice on human cadaver heads and anatomical phantoms, to avoid surgical complications. 


While cadaver heads offer the most realistic training setup, they are consumable (once drilled) and difficult to obtain. Anatomical phantoms are a great alternative but are expensive and also consumable. 
Moreover, feedback during such training is limited to subjective metrics such as Objective Structured Assessment of Technical Skills (OSATS), which are not comprehensive and limited in specificity \cite{laeeq2009pilot}. 

Computer simulations, on the other hand, offer several advantages that are difficult to replicate in a cadaver or phantom setup~\cite{george_de_2010}. First, virtual anatomies can be rapidly modeled based on actual patient scans, enabling patient-specific training and planning. Second, they are reusable and cost-effective. The initial setup is the only expensive part, allowing for extensive uses with low maintenance costs. Third, virtual surgeries can be replayed, making it possible to review the procedure for training and surgical competency assessment purposes. Lastly, these simulations can potentially produce invaluable data, which can be used for developing and validating downstream computer vision algorithms and artificial intelligence (AI) for various tasks. 

Developing such a simulation, or a ``simulation system'', requires extensive development to faithfully emulate a real surgical scenario. We use the phrase ``simulation system'' to emphasize that this system is more than a computer simulation, as it involves the integration of different hardware components alongside software algorithms. We present a Fully Immersive Virtual Reality System (\ours{}), which consists of an immersive volumetric drilling simulator that can simulate drilling on the patient Computer Tomography (CT) scans, a haptic device to control the virtual drill, a VR headset for immersive stereoscopic views, foot pedal interfaces to adjust the drill speed and change drill types, pupil detection and tracking hardware for gaze mapping, as well as an extensive data processing pipeline for recording and analyzing data for surgical skill assessment in real-time or offline. The entire system is based on open-source software and industry-standard data formats. To demonstrate the use cases, performance, and efficacy of \ours{}, we have conducted a pilot user study involving 3 attendants and 4 residents to perform mastoidectomy, a procedure in skull-base surgery. We present the preliminary quantitative results of these studies. We also present qualitative visualizations of \ours{}. 

\section{Related Work}
There has been considerable prior work in developing computer simulations for skull-base surgery. Early implementations of temporal bone dissection focus mainly on providing a simulation platform for training purposes~\cite{kuppersmith1997building, wiet2000virtual,agus2002mastoidectomy,pflesser2002volume}. Later, a non-patient specific skull drilling simulator with multi-point haptic rendering was presented in \cite{he2006bone}, increasing the simulation realism. In addition to 6~degrees of freedom (DOF) haptic rendering, another notable aspect of this work was the application of the local marching cubes algorithm for real-time updates and better visual fidelity. Further work by ~\cite{sorensen2009visible} and~\cite{wong2014comparison} greatly improved the visual fidelity of the anatomy and drilling operation.
Patient-specific simulation, which enables customized simulation and pre-operative planning was presented in~\cite{chan2016high}.
Recently, the OpenEar dataset provides cone-beam CT scans with micro-slicing to increase the resolution of simulated volumes \cite{sieber2021openear} for finer details. A commercial simulator called Voxel-man (Voxel-Man, Martinistr. 52, Hamburg, Germany) simulates various types of surgeries, including skull-base surgery, and it has been adopted in many hospitals for teaching purposes. 

The mentioned prior works focus on several different aspects for improving the simulation realism, to modeling patient-specific anatomy. On the other hand, \ours{} is a simulation system combining software \textit{and} hardware components for immersive training. Furthermore, \ours{} generates and records extensive data in an industry-standard format for quantitative skills assessment and for developing AI to push the boundaries of skull-base surgery. 




\section{Methods}

\begin{figure}[bt]
    \centering
    \includegraphics[width=\textwidth]{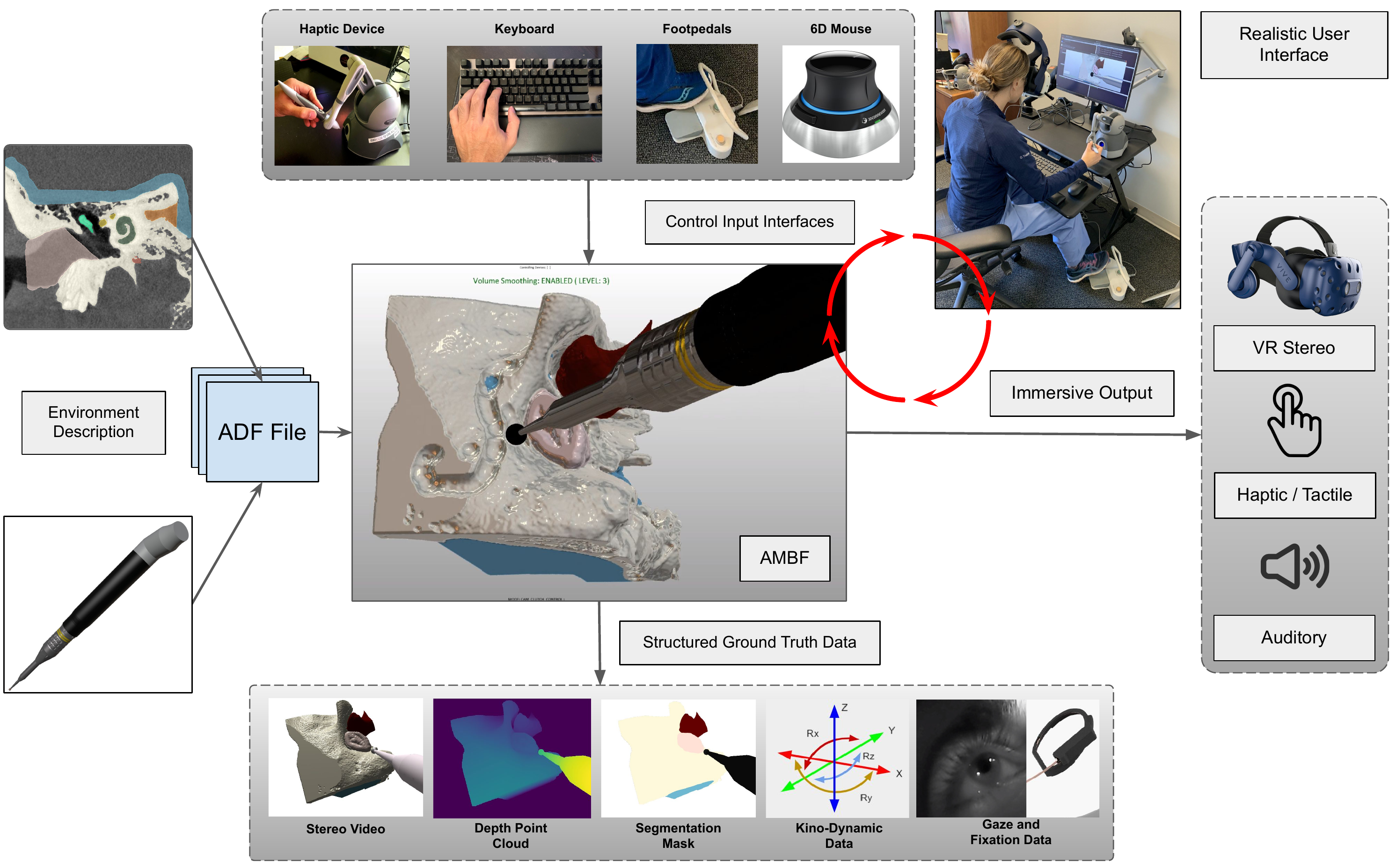}
    \caption{System architecture: \ours{} comprises of several components that work synergistically to provide a realistic user interface, visual fidelity, extensive data generation and recording features. The environment comprising of the anatomy, the drill and cameras are defined in ADF files which are loaded into AMBF. The drill, camera and the anatomy can be controlled/moved via a haptic device, keyboard or 6D mouse and a footpedal controls the drilling operation, similar to a real procedure. \ours{} generates immersive haptic, visual and auditory output for the operator. The ground truth data is generated and streamed asynchronously using Robot Operating System (ROS) middleware.}
    \label{fig: System Architecture} 
\end{figure}

\ours{} consists of several different hardware and software components, which are presented in Fig. \ref{fig: System Architecture}. We improve the realism of the system both in terms of the interface (Sec. \ref{sec: User Interaction and Interface}), and visual fidelity (Sec. \ref{sec: Visual Fidelity}). We also generate and record simulation data for downstream applications (Sec. \ref{sec: Data Generation and Management}).

\ours{} is developed on top of the Asynchronous Multi-Body Framework (AMBF)~\cite{munawar2019real,munawar2022virtual}. AMBF allows for the real-time simulation of complex robots and environments, both surgical and non-surgical, and provides interfaces for integrating multiple haptic devices alongside the physics simulation. All scene objects, dynamic objects, and haptic devices are defined in a front-end description format, called the AMBF Description Format (ADF). We also restructured and upgraded AMBF alongside \ours{} to create the necessary features required for its function. The upgrades include 1) a new multi-pass rendering pipeline, 2) the computation, and publishing of scene depth and video data in standardized Robot Operating System (ROS) formats, 3) the integration of volumetric model loading, drilling, and rendering, and 4) the modular integration with hardware and software components. The hardware components include head-mounted displays, foot pedal interfaces, pupil trackers, and 6 degrees of freedom (6D) mice as shown in Fig. \ref{fig: System Architecture}.

\ours{} is built upon AMBF's hierarchical plugin pipeline and includes multiple plugin types of different hierarchies. The core of \ours{} is implemented as a simulator plugin, while some optional components such as the virtual reality head-mounted display are implemented as object plugins. Details of various plugin types can be found in \cite{munawar2022virtual}.

\subsection{User Interaction and Interfaces}
\label{sec: User Interaction and Interface}
\subsubsection{Head Mounted Display}
The surgical field in skull-base surgeries is traditionally viewed via a high-resolution stereo microscope. We modeled our system to be similar to this setup by using a custom virtual reality (VR) headset on an off-the-shelf mount with custom 3D parts for interconnection. We developed a custom rendering pipeline for the VR/HMD interface that is based on OpenGL and Graphics Library Framework (GLFW). This allows us full control over the pre-processing and ``presentation'' aspect of VR. 

The camera view in the simulation can be changed by moving the haptic device (while pressing a button on the handle) or the 6D mouse as shown in Fig. \ref{fig: System Architecture}. In the future, we plan on incorporating external sensors for moving the simulation view by tracking the HMD pose.

\subsubsection{Modeling Drill Types and Bone Removal}
Various drill types are used for mastoidectomy procedures with different burr types and sizes. The burr types include cutting and diamond tips, and the common burr radii are 1, 2, 4, and 6 mm. The bone removal rate (BRR) consequently changes depending on the burr types and radii. Also, the bit shaft length is typically adjusted to be much longer for smaller burr sizes. To emulate this variation, we modeled the 4 drill burr radii (1, 2, 4, 6 mm) in \ours{}, and the BRR for each virtual drill is assigned based on surgeons' qualitative feedback. 
Further work is required to quantitatively calculate and emulate BRR based on real tissue. 

\begin{figure}[bt]
  \centering\begin{minipage}[t]{0.645\textwidth}
    \includegraphics[width=\textwidth]{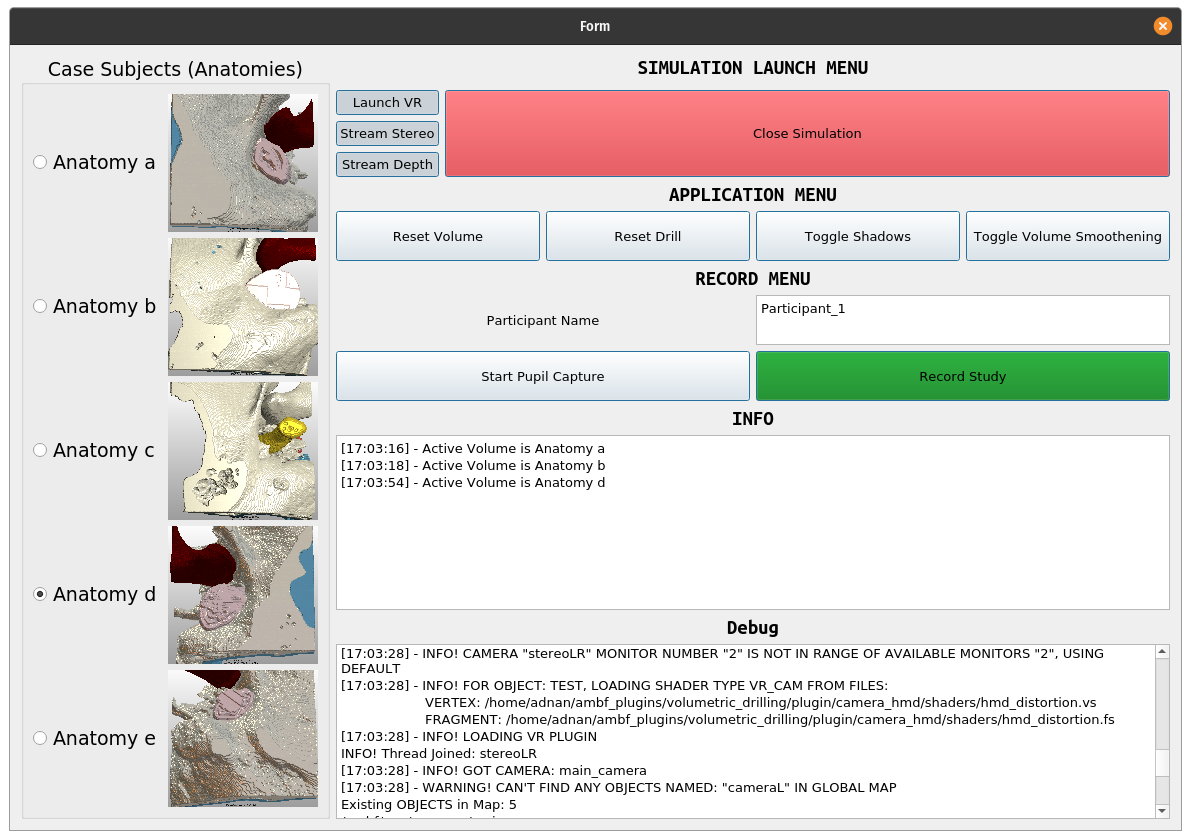}
    \caption{A intuitive GUI to launch, debug and test the simulation and its various components.}
    \label{fig: Simulation GUI}
  \end{minipage}
  \begin{minipage}[t]{0.34\textwidth}
    \includegraphics[width=\textwidth]{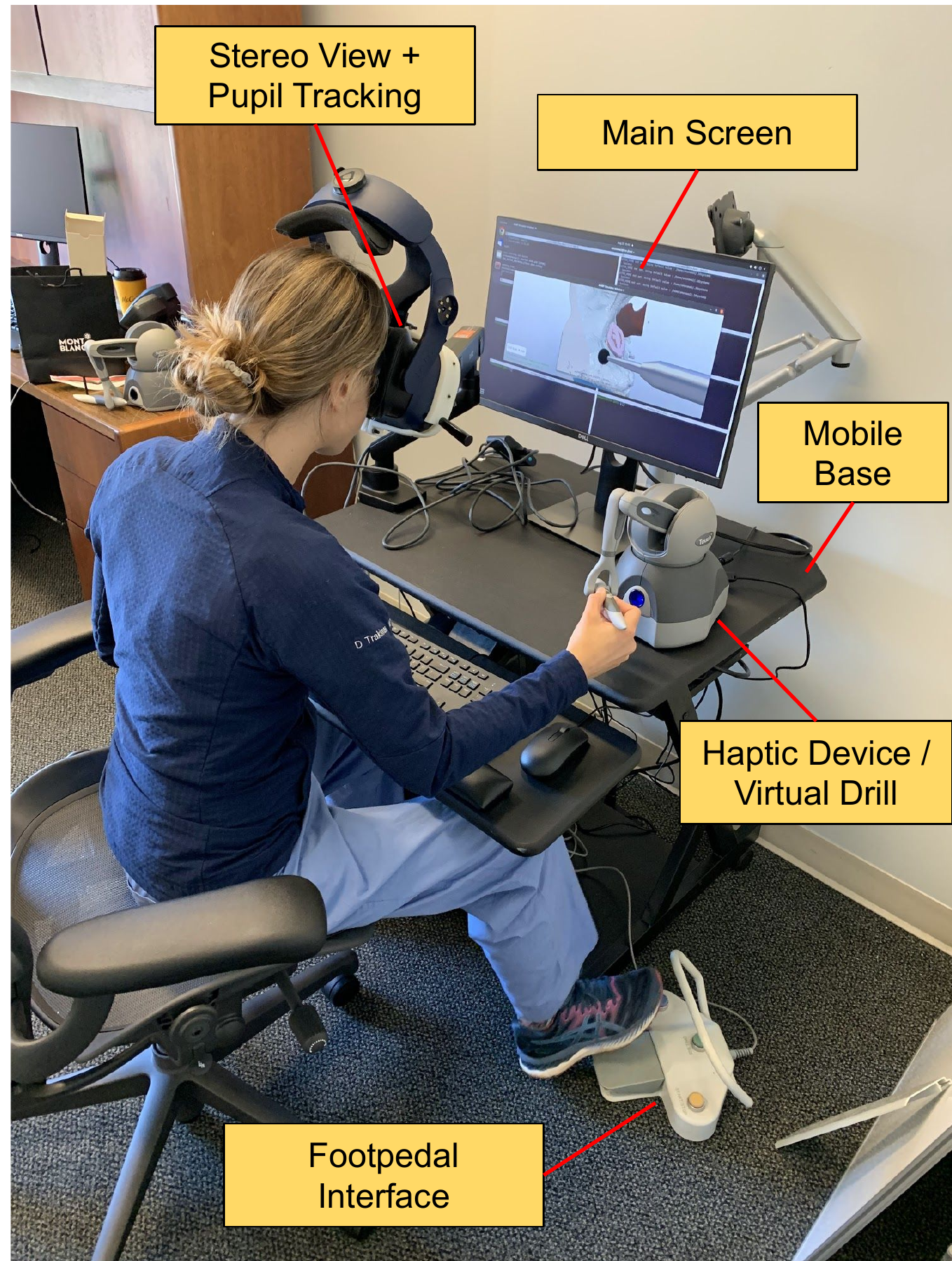}
    \caption{Labelled hardware setup}
    \label{fig: Setup at JHMI}
  \end{minipage}
\end{figure}

\subsubsection{Modeling Auditory Feedback}
In our previous work~\cite{munawar2022virtual}, we implemented haptic feedback to emulate the interaction of the drill with the anatomy as well as shaft collision detection. In our current work, we have incorporated audio feedback in addition to the haptic feedback to provide more realism.
An audio signal that mimics a drill's sound is played when the drill is powered on using the foot pedal interface and the cutting force alters the signal's pitch according to Eq. \ref{eq: audio pitch}.

\begin{equation}
    p = p_{max} - \|\vec{F}_{collision} \| / F_{max},
    \label{eq: audio pitch}
\end{equation}
where $p_{max}$ is the custom maximum audio pitch, $\vec{F}_{collision}$ is the force generated from the collision detection algorithm as described in \cite{munawar2022virtual}, and $F_{max}$ is maximum force threshold for the haptic device. Powering on the drill also adds vibratory feedback to the haptic device according to Eq. \ref{eq: haptic force},
\begin{equation}
    \vec{F}_{haptic} = \vec{F}_{collision} + \vec{1} \cdot A_{drill} \cdot \text{sin}(f \cdot t),
    \label{eq: haptic force}
\end{equation}
where $\vec{1}$ denotes a 3D vector of ones, $A_{drill}$ is the custom maximum vibration force magnitude, $f$ is the custom frequency, and $t$ is the system time. Furthermore, an informative text overlay has been added to inform/warn the operator in case the drill interacts with sensitive/forbidden anatomies.

\subsubsection{Graphical User Interface}

In our previous work, the drilling simulator had to be instantiated using the Linux command line which required some familiarity with BASH. While this is not an issue for research purposes, we had to deploy the system with unattended access for attending physicians and residents. Thus we developed an intuitive Graphical User Interface (GUI) that supports the multitude of options, as shown in Fig. \ref{fig: Simulation GUI}. A notable aspect of this GUI is that it instantiates as a standalone process and can execute, monitor and control the drilling simulator, pupil tracking software, and data recorder as separate processes.

\subsection{Visual Fidelity}
\label{sec: Visual Fidelity}
One of the more challenging aspects of creating a computer simulation that is attractive to attendants and residents is its visual fidelity. While the visual output from our previous work \cite{munawar2022virtual} was acceptable, the users (surgeons and residents) suggested specific shortcomings which included 1) unnatural artifacts/edges on the anatomy, 2) difficulty in perceiving depth and 3) mediocre illumination (and shading). \ours{} and the infrastructure behind it is based on OpenGL and thus we implemented the improvements as OpenGL shaders. While more work needs to be done, the users were satisfied with the updates. Each implementation for improving the visual fidelity is discussed below.

\begin{figure}[!bt]
  \centering
  \begin{minipage}[t]{0.49\textwidth}
    \includegraphics[width=\textwidth]{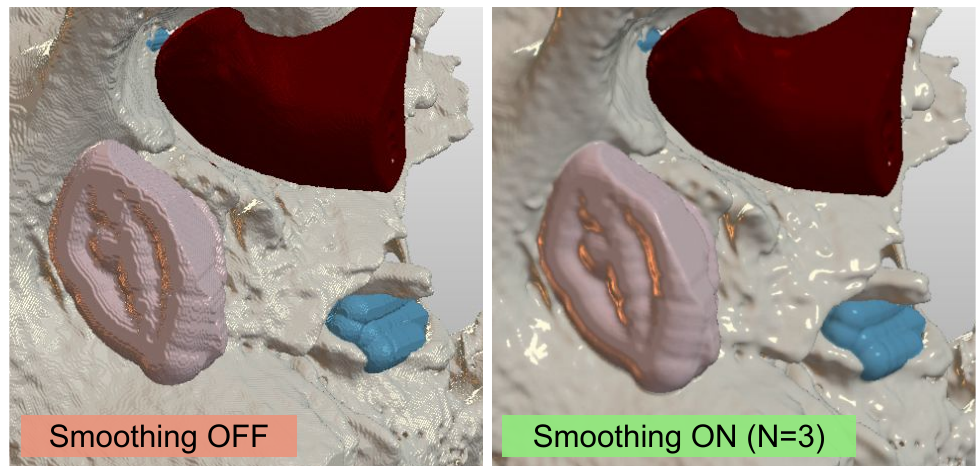}
    \caption{Comparison between online volume smoothing turned on and off}
    \label{fig: Volume Smoothing} 
  \end{minipage}
  \begin{minipage}[t]{0.49\textwidth}
   \includegraphics[width=\textwidth]{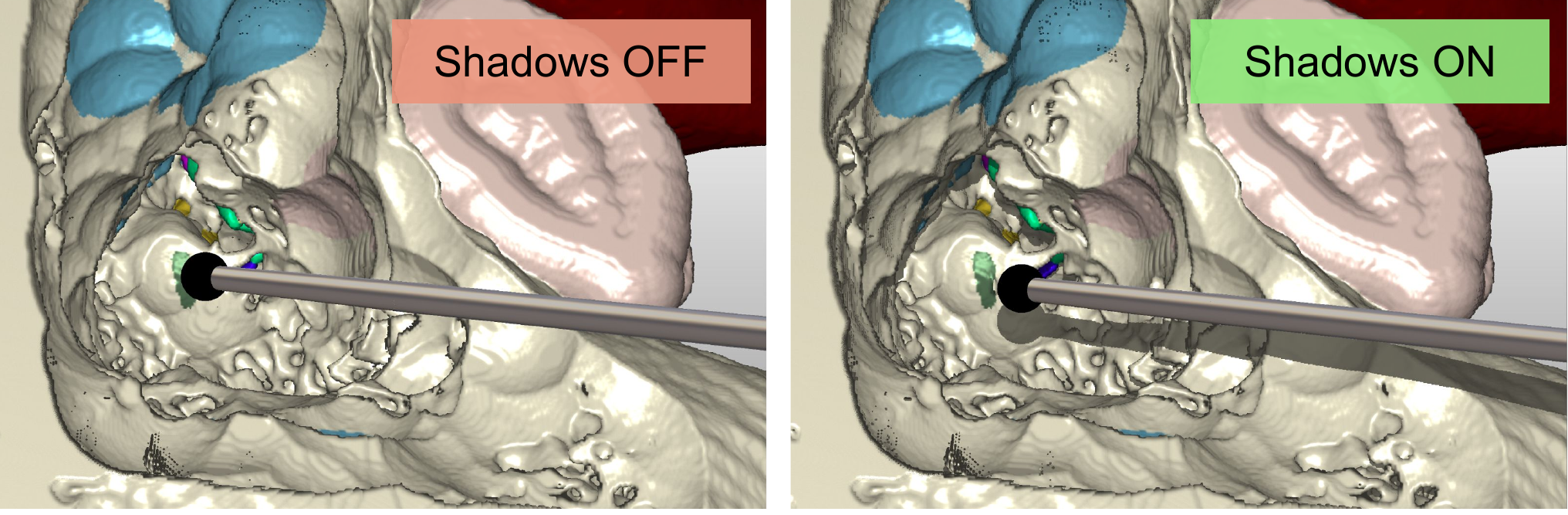}
    \caption{Comparison between shadowing, to improve depth perception, turned on and off}
    \label{fig: Shadowing} 
  \end{minipage}
\end{figure}

\subsubsection{Online Smoothing of the Iso-surface}
 The volumetric scans that we used for \ours{} comprise the middle + inner ear region and are roughly $100~\text{mm}^3$ in dimensions with an approximate resolution of $500^3$ voxels. This resolution, while suitable for discerning and segmenting the smaller anatomies within, is less than ideal for rendering without smoothening the surface as shown in Fig. \ref{fig: Volume Smoothing}. For better rendering quality, prior work generally constructs a static geometric mesh from the scan and applies surface smoothening on the mesh.  

Computing the surface mesh for the volumetric data (using algorithms such as marching cubes) is usually performed offline as it is a moderately expensive operation. This is not a problem for a constant grid of data and the advantage of creating a geometric mesh is that the resulting mesh can be further processed, i.e., surface smoothening. However, in our case, the volumetric data is modifiable via simulated drilling operations, and thus recreating the surface mesh is not ideal. Instead, we implement iso-surface smoothing via a volumetric ray casting approach implemented on the GPU (Alg. \ref{algo: volume smoothing}).

\begin{algorithm}
\scriptsize
\caption{Online Volume Smoothing}
\label{algo: volume smoothing}
\begin{algorithmic}[1]
\State{$\delta \vec{p} = \vec{1} * (N-1) / 2.0$} \Comment{$N :=$ Sample Threshold $\in \mathbb{R}$ }
\State{$\vec{\phi} = \vec{1} / [c_x, c_y, c_z]'$} \Comment{$c_x, c_y, c_z := $ Number of Voxels along X, Y and Z}
\State{$\vec{p}_{iso} := Raycast(\vec{v})$} \Comment{$\vec{v}:=$ View Ray. $\vec{p}_{iso }$ refined by interval bisection }
\State{$\vec{\eta} = \vec{0}$} \Comment{Smoothed Normal}
\For{$x \lt N$} 
    \For{$y \lt N$}
        \For{$z \lt N$}
            \State{$\vec{p}_{offset} = [x, y, z]'$}
            \State{$\vec{p}_{sample} = \vec{p}_{iso} + ( \vec{p}_{offset} - \delta \vec{p}) * \vec{\phi}$ } 
            \State{$\vec{\eta} = \vec{\eta} + \nabla(\vec{p}_{sample})$} \Comment{$\nabla :=$ Gradient from Central Difference}
        \EndFor
    \EndFor
\EndFor
\State{$\vec{\eta} = normalize(\vec{\eta})$}
\end{algorithmic}
\end{algorithm}

 The resulting normal $\vec{\eta}$ from the above algorithm is then used for lighting calculations and this simple approach produces noticeably improved visualization of the anatomy. Depending upon the sample threshold $N$, the rendering frame rate (FPS) is reduced. In our experience, running FIVRS on a computer with AMD Ryzen 5 3600, 32 GB DDR4 RAM, and an NVidia GTX 1080, the FPS dropped from the intended 120 Hz to 102 Hz for $N=3$. Please see the supplementary video for a better comparison between the visual outputs.

\subsubsection{Shadow Mapping}
We implemented shadow mapping (both external object and self-induced) to the rendered volume to improve depth perception (Fig. \ref{fig: Shadowing}). The improvement is especially noticeable when viewed stereoscopically (i.e., via HMD). Shadow mapping on volumetric data rendered via ray-casting shares the same principles as geometric meshes, however, the implementation is a bit different as the volume is not rendered using primitives (triangles or quads).


\subsubsection{Lightmaps}
Physically based rendering (PBR) is the state-of-the-art technique for achieving visual realism in computer simulations. This technique uses the reflectance equation instead of the simpler Blinn-Phong shading model to compute the fragment/pixel intensity (color). The reflectance equation produces spectacular visualizations, however, it is considerably more computationally expensive.  We instead improve the visual fidelity by incorporating a computationally and algorithmically simpler lighting model based on pre-baked lightmaps. We plan to integrate PBR for AMBF and \ours{} for real-time rendering in the future.

\subsection{Data Generation and Management}
\label{sec: Data Generation and Management}

\begin{figure}[ht]
    \centering
    \includegraphics[width=\textwidth]{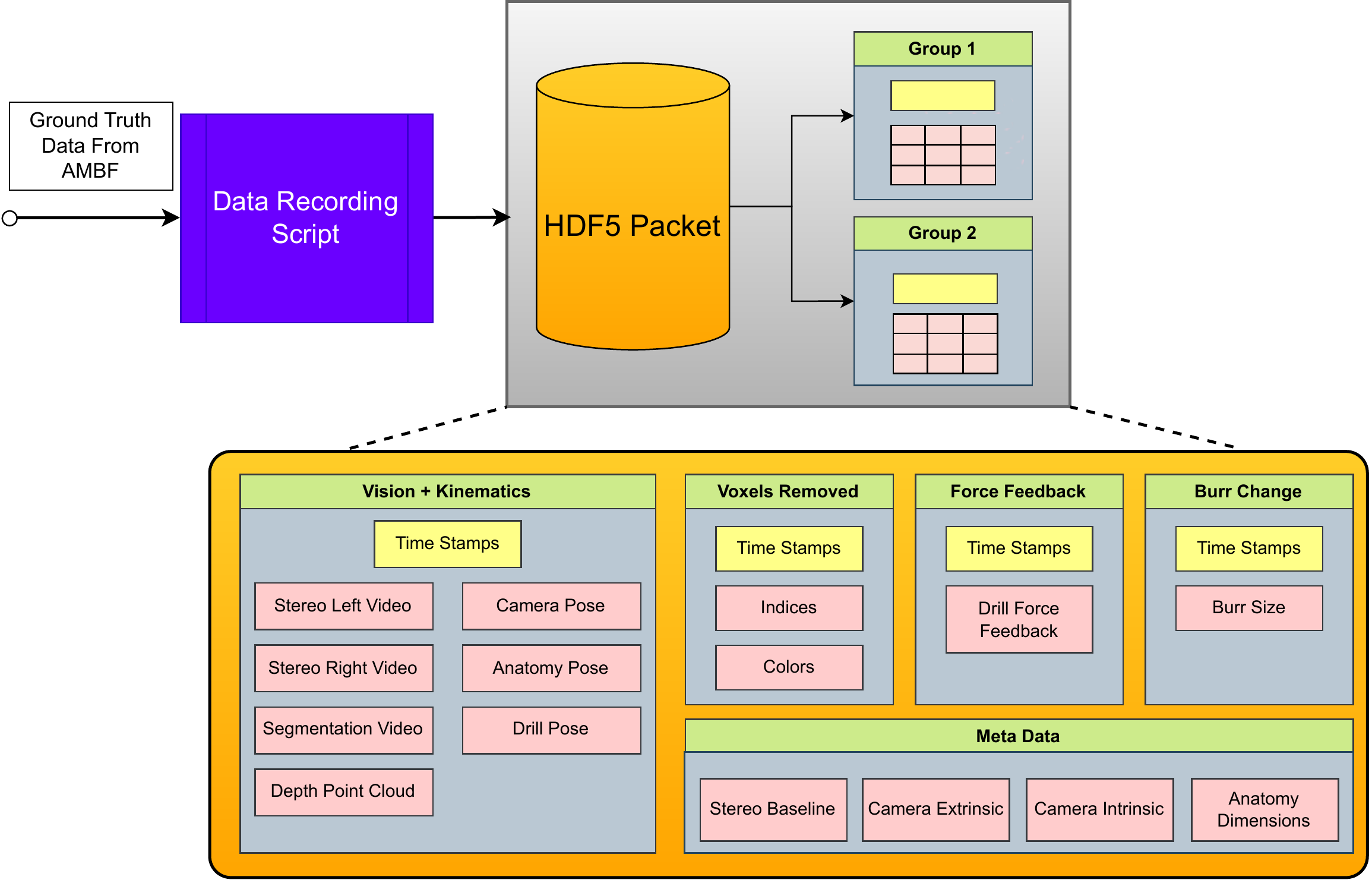}
    \caption{The asynchronous data from \ours{} is captured by a Data Recording script which synchronizes and writes this data as HDF5 files. The data is split into various ``groups'' with appropriate meta-data. The ``voxels removed'' group contains the 3D indices of the removed voxels, the time of removal, and their associated labels (colors). The ``force feedback'' is the force applied to the haptics device, and ``burr change'' is the change of burr sizes and time stamps. The ``vision + kinematics'' and ``meta data'' are self-explanatory.}
    \label{fig: HDF5 Payload} 
\end{figure}
\subsubsection{Importing Anatomical Scans}
Volumetric data from CT and MRI scanners is often output in the Digital Imaging and Communication (DICOM) format. In prior work, we collected a large data set of patient CT scans and manually segmented the sub-cranial anatomies using 3D Slicer \cite{ding2021automated}. Manual segmentation is an effort-intensive task and automated techniques such as \cite{you2022simcvd} can be potentially beneficial. The resulting scans were then saved as Segmented Nearly Raw Raster Data (Segmented NRRD ``.seg.nrrd'') which is a superset of the ``NRRD'' format used by 3D Slicer. \ours{} can render either intensity-based volumetric data (NRRD), or constant intensity pre-segmented volumes. The latter format is extracted from the 3D Slicer specific ``seg.nrrd'' format in which each segment is either one hot encoded into a separate grid layer or labeled incrementally (1,2,3 ... for different segments). We created a program that converts the data in ``.seg.nrrd'' to an array of images in either JPEG or PNG format as these raw images are easier to be incorporated into \ours{}. The spatial data from the ``.seg.nrrd'' files are encoded separately using ADF files. 

\subsubsection{Ground Truth Vision Data}
For \ours{}, we render 4 simulated cameras and each of these cameras requires multi-pass rendering. The four cameras include a single-frame left-right stereo for viewing on the HMD, separate left and right stereo cameras with different baselines (for recording and evaluation purposes), and a segmentation camera. For the single stereo HMD camera, multiple rendering passes include rendering for the left and right halves of the viewport, applying the inverse geometric and chromatic distortion filters, and finally rendering to the screen. For the segmentation camera, the rendering passes include computing the segmentation masks and then separately the depth point cloud. In addition to multi-pass rendering, three of these cameras (other than the single stereo camera) stream their video, and the depth point cloud in the case of the segmentation camera, over ROS. The generated data can thus be used to train deep-learning-based algorithms for depth estimation~\cite{li2021revisiting,li2023temporally}, motion tracking~\cite{li2022tatoo} and segmentation~\cite{shvets2018automatic}.

Computing the data mentioned above resulted in reduced frame rates. In this work, we have optimized AMBF's rendering and depth computation pipeline to alleviate the computational load. First, we implemented an optional presentation-free mode for each camera, which allows any camera to only stream its video and not present/render to the screen. Second, we reorganized the depth computation algorithm from our previous work \cite{munawar2022virtual} to take place entirely on the GPU by leveraging floating point textures. These optimizations increase the overall frame rate by $15$ FPS. 

\subsubsection{Gaze Mapping}
Pupil tracking and gaze mapping can demonstrate what specific structures surgeons focus on. This data may further predict the operator's workload. To record this data, we incorporate the pupil eye tracking hardware (Pupil Labs GmbH, Sanderstr. 28, 12047, Berlin) into our HMD. The pupil tracking hardware outputs videos for each eye and associated time stamps which we post-process to compute the operator's gaze, fixation, and blinks.

\subsubsection{Data Recording Pipeline}
One of the notable features of \ours{} is its ability to record multi-modal data, including videos (from stereo cameras), real-time depth point cloud, segmentation mask, as well as kinematic and dynamics of scene objects (cameras, drills, anatomy). 
While ROS bags are convenient and often used to store data, they are not optimized for storage. As an example, the file size of a virtual surgery recorded at $\approx 10~\text{FPS}$ and $640\times480$ resolution for $60~\text{seconds}$ occupies $\approx 15~\text{GB}$ of storage. 
For a single extended mastoidectomy session that may last 20 minutes, the resulting file size of $\approx 300~\text{GB}$ is almost impractical to manage. 
Instead, we leverage the HDF5 data format with compression to reduce the footprint to $\approx 100~\text{MB}$. Furthermore, for each individual study, we split the recorded data into multiple files based on a user-defined batch size, thus avoiding unexpected loss of data. The pupil-tracking data is stored separately after synchronizing the timestamps with the HDF5 files. The data in each HDF5 file is organized as shown in Fig. \ref{fig: HDF5 Payload}.


\begin{figure}[ht]
    \centering
    \includegraphics[width=\textwidth]{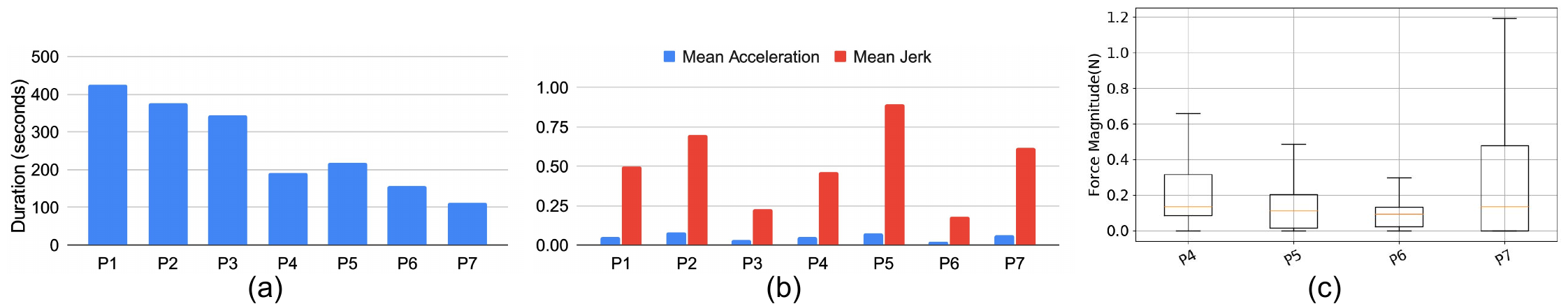}
    \caption{Results for performing cortical mastoidectomy by participants P1-P7. b) Acceleration and Jerk are reported in Simulation Units. c) Force data for only 4 participants was collected (P4-P7)}
    \label{fig: Evaluation Metrics} 
\end{figure}

\section{Experiments and Results}
We conducted a pilot study with 7 participants including 3 attending surgeons (P4, P5, P7), 3 residents (P1, P2, P3), and 1 medical student (P6), who performed a cortical mastoidectomy on a setup shown in Fig. \ref{fig: Setup at JHMI}. The participant ages ranged from 23-37 years and included 4 males and 3 females. The purpose of the study was to perform an initial validation of the system and the recorded data rather than characterize the performance of the participants. The user study was approved by our institutional review board under IRB00264318 and was conducted at both the medical campus as well as the engineering campus. Two of the three residents had significant prior experience with the procedure. To account for the learning curve and to establish familiarity with \ours{}, each participant interacted with the system and operated on multiple anatomies for at least 30 minutes. After the training sessions, the study participants operated on different anatomies, and the data presented in Fig. \ref{fig: HDF5 Payload} along with pupil tracking was recorded. 

\begin{figure}[bt]
    \centering
    \includegraphics[width=\textwidth]{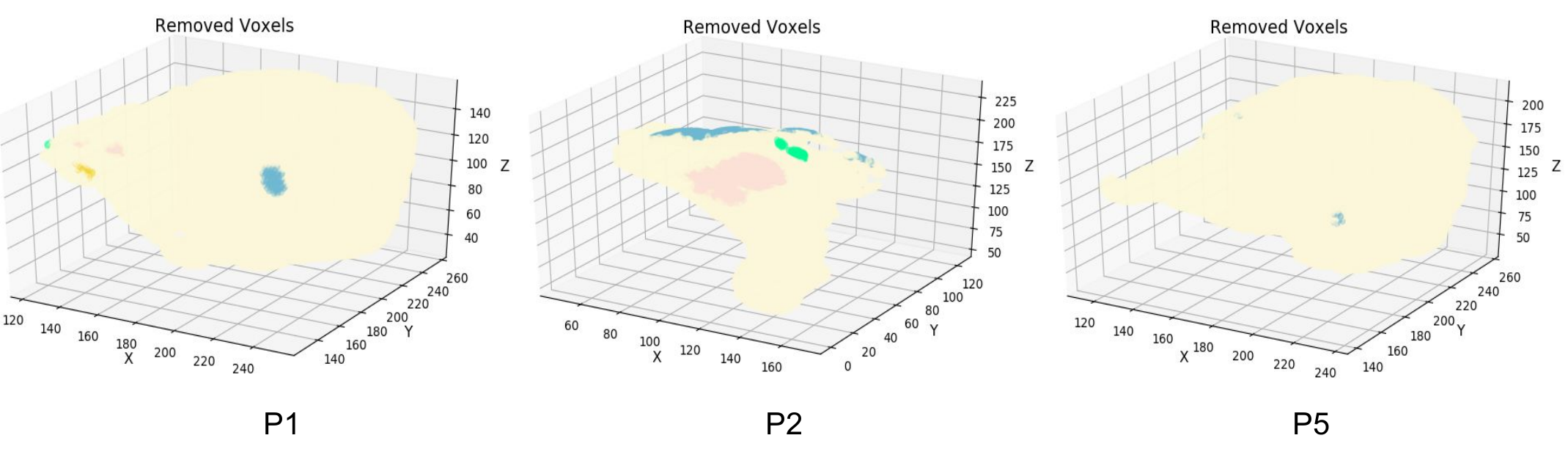}
    \caption{Visualization of volume (voxels) removed. Colors other than bone (pastel) represent unintended removal of sensitive/critical structures}
    \label{fig: 3D Volume Removal} 
\end{figure}

We present some results based on common anatomy. Because of the small participant count and lack of a concerted effort for diverse demographics, the study is primarily limited to only showcase the utility of the system and the recorded data. Fig. \ref{fig: Evaluation Metrics} shows the temporal, kinematic, and dynamics data for each participant. Fig. \ref{fig: 3D Volume Removal} shows an interesting metric which is the volume removed. This data will be invaluable for machine learning applications and other tasks such as predicting the volume to be removed for a given anatomy.



\section{Conclusion and Future Work}
We present \ours{}, a fully immersive VR system for skull-base surgery. In addition to providing a realistic ``simulation system'' for skull surgery, \ours{} offers an extensive data generation and recording interface using an industry-standard format. \ours{} is lightweight, modular, and extensible. The infrastructure is already being used for numerous research applications including simulation of other medical procedures, surgical navigation \cite{shu2022twin}, and enforcing virtual fixtures~\cite{ishida2023improving} for robot control with simulation in the loop. 

In the future, we plan to continue collecting data from surgeons to create an extensive data set for several applications ranging from developing a quantitative metric for surgical skills evaluation, improving surgical workflow by presenting the predicted anatomy to be removed to the surgeons, and integrating with robots (such as the ENT Robot from Galen Robotics, 1100 Wicomico St., Baltimore, MD) for safer cooperative control with simulation-based virtual fixtures.

\bmhead{Acknowledgments} This work was supported by NSF OISE-1927354 and OISE-1927275, NIDCD K08 Grant DC019708, a grant from Galen Robotics, and an agreement between Johns Hopkins University and the Multi-Scale Medical Robotics Centre, Ltd.

\section*{Declarations}

\bmhead{Conflict of interest} Russell Taylor and JHU may be entitled to royalty payments related to technology discussed in this paper, and Dr. Taylor has received or may receive some portion of these royalties. Also, Dr. Taylor is a paid consultant to and owns equity in Galen Robotics, Inc.  These arrangements have been reviewed and approved by JHU in accordance with its conflict of interest policy.  

\bibliography{sn-bibliography}


\end{document}